\documentclass[letterpaper, 10 pt, conference]{ieeeconf}
\IEEEoverridecommandlockouts
\usepackage{cite}
\usepackage{amsmath,amssymb,amsfonts}
\usepackage{algorithmic}
\usepackage{graphicx}
\usepackage{textcomp}
\usepackage{xcolor}
\def\BibTeX{{\rm B\kern-.05em{\sc i\kern-.025em b}\kern-.08em
    T\kern-.1667em\lower.7ex\hbox{E}\kern-.125emX}}
\begin{document}

\title{Reducing Tactile Sim2Real Domain Gaps via Deep Texture Generation Networks\\
}

%\title{Unsupervised Deep Domain Adaptation for Tactile Features\\}

\author{Tudor Jianu, Daniel Fernandes Gomes and Shan Luo% <-this % stops a space
%\thanks{*This work was supported by the EPSRC project ``ViTac: Visual-Tactile Synergy for Handling Flexible Materials" (EP/T033517/1).}% <-this % stops a space
\thanks{All the authors are with Department of Computer Science, University of Liverpool, Liverpool L69 3BX, United Kingdom. Emails: \tt\{t.jianu, danfergo, shan.luo\}@liverpool.ac.uk}
}

\maketitle

\begin{abstract}
%1. [Background]
%Daniel&Shan
Recently simulation methods have been developed for optical tactile sensors to enable the Sim2Real learning, i.e., firstly training models in simulation before deploying them on the real robot. However, some artefacts in the real objects are unpredictable, such as imperfections caused by fabrication processes, or scratches by the natural wear and tear, and thus cannot be represented in the simulation, resulting in a significant gap between the simulated and real tactile images. To address this Sim2Real gap, we propose a novel texture generation network that maps the simulated images into photorealistic tactile images that resemble a real sensor contacting a real imperfect object. Each simulated tactile image is first divided into two types of regions: areas that are in contact with the object and areas that are not. The former is applied with generated textures learned from real textures in the real tactile images, whereas the latter maintains its appearance as when the sensor is not in contact with any object. This makes sure that the artefacts are only applied to the deformed regions of the sensor. Our extensive experiments show that the proposed texture generation network can generate these realistic artefacts on the deformed regions of the sensor, while avoiding leaking the textures into areas of no contact. Quantitative experiments further reveal that when using the  adapted images generated by our proposed network for a Sim2Real classification task, the drop in accuracy caused by the Sim2Real gap is reduced from 38.43\% to merely 0.81\%. As such, this work has the potential to accelerate the Sim2Real learning for robotic tasks requiring tactile sensing.

\end{abstract}

%\begin{IEEEkeywords}
%Tactile sensing, simulation, Sim2Real learning, generative models
%\end{IEEEkeywords}

\section{Introduction}
% [Background]
Humans have a native capability to manipulate objects without much effort. We use different sensing modalities, e.g., visual, auditory and tactile sensing, to perceive the object properties and manipulate them in space. Among these sensing modalities, tactile sensing is not affected by changes of light conditions and occlusions of hands as vision, or influenced by the noise of the ambient environment. It can provide us rich information of the object in hand, e.g., the texture, temperature, shape and pose of the object.

To equip the robot with similar tactile sensing capabilities, various tactile sensors have been proposed for robots in the past decades to imitate the human skin~\cite{TactileSensingRobotHandsSurvey,Directionstowardeffective,luo2017robotic}. Traditional approaches aimed at providing the robot with force information at a contact point~\cite{tactileSensing}. In order to provide the robot with more information on the contact, camera-based optical tactile sensors have been proposed and the GelSight sensor is one of them. The GelSight sensor captures high resolution geometric information of the object it interacts with, thus having the capability to aid the manipulation task~\cite{gomes2021generation}. It uses a camera to capture the deformation of a soft elastomer, using illumination sources from different directions~\cite{gomes2021generation}. It also has a few variants of different morphologies and camera/light configurations such as GelTip~\cite{gomes2020geltip} and GelSlim~\cite{donlon2018gelslim}. 

\begin{figure}[t]
\centering
\centerline{\includegraphics[scale=0.31]{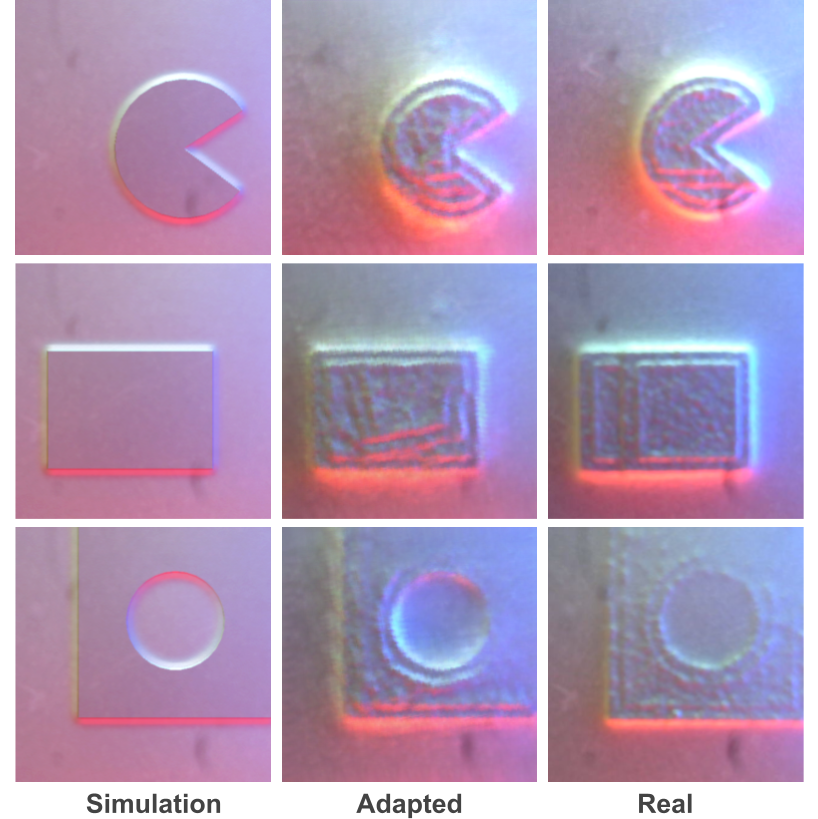}}
\caption{The adapted tactile images (middle) are generated from the simulation tactile images (left) using our proposed texture generation network. Compared to the corresponding real images (right), it can be observed that the
contact regions in the adapted tactile images have similar textures to ones in the real images, with the uncontacted areas free from the textures.}
\label{fig:cover}
% \caption{The adapted images compared to the real images. It can be observed that the model is able to generate different deformations of the object resulting in a more robust adaptation.}
\label{overview}
\end{figure}

Due to the use of a soft elastomer on the top of the sensor to interact with objects, similar to many other tactile sensors, the camera based tactile sensors are fragile and suffer from wear and tear. To mitigate the damage to the sensor and save time for training on a real robot, the robot can be trained in a simulated environment first with simulated tactile sensors, before deploying the trained model in a real environment. To this end, simulation models of the GelSight sensor have been proposed~\cite{gomes2019gelsight,gomes2021generation}.

%2. Unmet needs \\
However, the gap between the simulated tactile environment and the real world is still large, which may perturb the model and greatly impact its performance. As discussed in~\cite{gomes2021generation}, the imperfections of reality such as the scratches and other object deformation are what helps the trained model distinguish between objects via touch sensing. In contrast, in the simulation, those are not present, thus creating a gap between the two domains. One of the methods to diminish such gaps is to make the simulation as real as possible. In order to do so, the introduction of noise either in the form of textures and other methods such as adding Gaussian noise or domain randomisation can be adopted~\cite{tobin2017domain,domainRandomization2}. But its main issue is that the probability distribution of the imperfections in the reality domain has a long tail: although the probability of encountering a novel imperfection is small, it will eventually happen. In the context of robotic manipulation tasks, this can become a potential dangerous situation or damage the sensors~\cite{gomes2021generation}.

To address this challenge, we propose a novel texture generation network to reduce the domain gap from simulated tactile images to the real tactile images for camera-based optical tactile sensing.
%4. Rationale \\
In the proposed network, different regions of the simulated tactile image are adapted differently: the areas in contact with the object are applied with the generated textures from real tactile images, whereas regions without a contact maintain their appearance as when the sensor is not in contact with any object.
%5. Experiment results \\
We have conducted extensive experiments to evaluate the proposed method using a dataset of real and simulated tactile images from a GelSight tactile sensor. The experiments show that the proposed method can generate realistic artefacts on the deformed regions of the sensor, while avoiding leaking the textures into ones without a contact, as shown in Fig.~\ref{fig:cover}. In comparing the resulted tactile images with real tactile images, it achieved a low Mean Absolute Error (MAE) of 10.53\% on average and a similarity of 0.751 in the Structural Similarity Index (SSIM) metric. Beyond that, the experiments show that when using the adapted images generated by our proposed network for Sim2Real transfer of a learnt model for a classification task, the drop accuracy caused by the Sim2Real gap is reduced from $38.43\%$ to merely $0.81\%$.
%6. Impact or Importance of your work scientifically \\
As such, this work has the potential to accelerate the Sim2Real learning for robotic tasks with tactile sensing.

\section{Related Works}

\subsection{Optical tactile sensors}
Optical tactile sensors that use a camera underneath a soft elastomer layer are one highly practical method for providing robots the sense of touch, and thus a variety of sensor designs have been proposed. Currently, these can be grouped in two main families: marker-based, represented by TacTip sensors~\cite{TacTipFamily}, and image-based, represented by GelSight  sensors~\cite{dong2017improved}. %While the first works by tracking a grid of markers printed on the inner side of  the sensor’s membrane, the second considers the  higher resolution raw  image for photometric analysis. 
In this paper, we focus on GelSight sensors as they are better suited to capture the fine textures introduced by manufacturing defects or wear and tear. The \textit{GelSight} working principle was proposed in~\cite{RetrographicSensing} as a method for reconstructing the texture and shape of contacted objects\cite{cao2020spatio,luo2018vitac,lee2019touching}. To that purpose, light sources are placed from opposite angles next to a transparent elastomer that is coated with an opaque reflective paint, resulting in three different shaded images of the in-contact object texture. A direct mapping between the observed image pixel intensities and the elastomer surface orientation can then be found to create a lookup table, enabling the surface to be reconstructed using photometric stereo. Since its initial proposal, new designs have been proposed that aim at reducing the size~\cite{dong2017improved, donlon2018gelslim}, improving its sensing capability~\cite{donlon2018gelslim, dong2017improved} or providing curved finger-shaped surface for improved robotic dexterity~\cite{BlocksWorldOfTouch, softRoundGelSight, cao2021touchroller}. %, and much research has been produced using such sensors.
However, optical tactile sensors are still brittle and extensive experimentation with them often results in their sensing membrane being damaged. 

\subsection{Simulation of tactile sensors}
It is desirable to develop and test robot agents initially within a simulator before their deployment in the real environment as running experiments with real hardware is time consuming and damage prone. To this end, a variety of methods have been proposed to simulate different tactile sensors. We proposed to simulate GelSight sensors in~\cite{gomes2019gelsight, gomes2021generation}, by considering close-up depth-maps extracted from simulators and using the Phong illumination model for rendering the RGB tactile images. %In~\cite{tacto}, the direct usage of OpenGL rendering engine has been proposed and~\cite{phisicsBasedLight} physics based light modelling has been considered. 
However, despite the efforts in making the simulations as realistic as possible, some artefacts such as the textures that are not represented in the simulated object model and the scratches resulting from wear and tear contribute largely to the gap between the simulated and real images. They often hinder transferring the models trained on simulated data to the real robots (i.e., Sim2Real learning). Therefore, the gap itself must be addressed.

\subsection{Reducing the Sim2Real gap for tactile sensing}

A common approach to address the gap between training and test data is to augment the training data such that the test data becomes one particular subset of the whole augmented training data, i.e., Domain Randomisation. For Real2Real computer vision tasks, this augmentation is commonly performed at the image level, by applying random colour or geometric transformations to the images. When the training images are collected in simulation for Sim2Real learning, one other form of augmentation is to directly augment the simulation data, by randomising object colours, scene illumination or the environment physics~\cite{tobin2017domain}. As colours and illumination are constant for tactile images collected from the same tactile sensor, in~\cite{gomes2021generation} we experiment with augmenting the synthetic dataset by perturbing the in-contact object shapes using simple texture maps that resemble the artefacts mostly contributing to the gap observed in our dataset: the textures introduced in the 3D printing of our real reference object dataset. The method was proved to be a more effective augmentation schema than image-based augmentations. However, target domain agnostic randomisation is costly as it often requires running the same simulation in a great number of times, to capture all the dimensions' variations. Thus, in this paper we address the Sim2Real gap from a Domain Adaptation perspective and propose a network that adapts the simulated images into photo realistic counterparts. %\tudor{Furthermore, as we are interested in capturing the artefacts of the real contacted objects that are not modelled in the simulation, different regions  of the  simulated  tactile  image  are  adapted  differently: areas of the image that correspond to the in-contact object  will  be  applied  with  the generated  textures  from  real  tactile  images,  whereas  regions corresponding to the background elastomer will  maintains  its  appearance  as  when  the sensor  is  not  in  contact  with  any  object.}
Domain adaptation has successfully been applied to vision-based tasks~\cite{cyCADA, photorealismEnhancement}, however, it has not been studied in the context of tactile sensing yet.

\section{Methodology}

\begin{figure}[t]
\centerline{\includegraphics[scale=0.45]{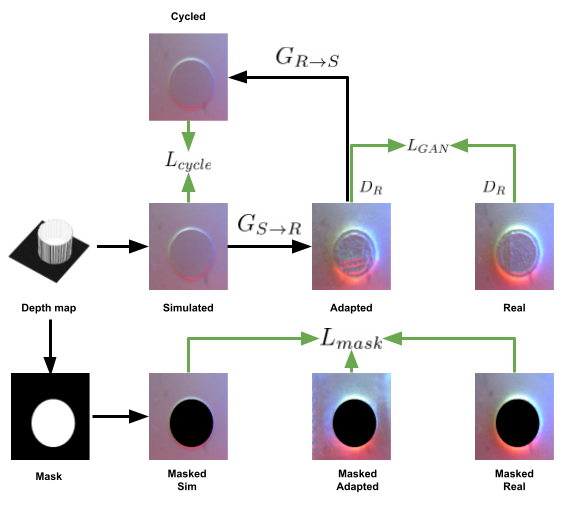}}
\caption{The proposed texture generation network. Starting from the \textbf{Depth map} captured in the simulator, the \textbf{Simulated} tactile image and \textbf{Mask} are generated using \cite{gomes2021generation} and simple truncation of the depth map, respectively. The simulated image is then mapped to the \textbf{Adapted} target $R$ through $G_{S \rightarrow R}$. A discriminator $D_R$ then classifies the generated tactile image $G_{S \rightarrow R}($\textbf{Simulated}$)$ thus giving the adversarial loss $L_{GAN}$. The image is then cycled back to the simulated domain through $G_{R \rightarrow S}$, which then gives the cycle consistency loss ($L_{cycle}$). The mask of the in-contact area is used to cover the contact zones of the \textbf{Simulated}, \textbf{Adapted} and \textbf{Real images}, thus allowing us to constrain the background of the tactile image, while allowing the model to alter the contact zone with textures, resulting in $L_{mask}$.}
% try to use the words in the picture. you are describing the method, using the picture. 
\label{framework}
\end{figure}

\subsection{Problem description}
%1. Ideal \\
%Robotics training often require an abundance of data in order to be able to infer complex patterns and to succeed in a given task. Giving the large scale data required for such a task, a common approach is to simulate data which would otherwise be impractical to collect in the real world. By simulating the real environment, a model benefits the advantage of being trained on an abundant amount of data that can be obtained at a lower cost in relation to both monetary and time resources. 
%2. Reality\\
In this paper, we address the domain gaps between the simulation tactile images and real tactile images, for the first time, that impede the ability of transferring a trained model in simulation to reality. The resulting factor that contributes to the domain gaps is represented by the textural artefacts~\cite{gomes2021generation}. Those artefacts are not limited to the production phase such as manufacturing defects and surface textures brought by the finishing, but can be created when an object is repeatedly interacted with thus being continuously deformed, i.e., wear and tear. Failure to consider the artefacts when training the robot agents can lead to an improper manipulation of a given object due to miss-classification and ultimately result in a possible damage to the robot.

%However, the simulation often differs from the reality and result in domain gaps that impede the ability of a model to be transferred seamlessly to reality after it has been trained in a simulation. In the case of tactile images, the resulting factor that contributes to the domain gap is represented by the textural artefacts that resulted in the process of 3D printing the given objects \cite{gelsightMain}. Those artefacts are not limited to the production phase but can be created when and object is repeatedly interacted with thus being continuously deformed. Those deformation can lead to an improper manipulation of a given object due to miss-classification and ultimately result in a possibly dangerous situation. Additional textural augmentation leaks, can have greater impediments when harder tasks such as segmentation with the purpose of localisation are employed, as improper localisation of the contact region would yield an incorrect assessment of the situation.  

%3. Consequences\\
To this end, we aim at addressing the gap between simulated and real tactile images to diminish the risk of such situation and propose to learn the artefacts on object surfaces so as to mitigate the drop in performance in Sim2Real learning. It is challenging as texture artefacts should be applied only to the contact regions of the tactile sensors with the rest unaffected, as textural augmentation leaks to the untouched areas may lead to fake positive detection of contacts. In order to do so, we propose a novel texture generation network for applying textures to the contact surfaces in the simulation tactile images.

\subsection{The texture generation network}

As shown in Fig.~\ref{framework}, our proposed texture generation network has two generators: one generating a tactile image with textures $\hat{X}_A$ (i.e., an adapted tactile image) from a simulation tactile image $X_S$, i.e., $G_{S\rightarrow R}$; the other generating tactile images in the simulation domain from the adapted tactile image, i.e., $G_{R\rightarrow S}$. Two discriminators are responsible for distinguishing the real image from the generated one created by the generator in each domain: Discriminator $D_R$, aims at distinguishing $X_R$ from $G_{S\rightarrow R}(X_S)$ while the discriminator $D_S$ aims at distinguishing between $X_S$ and $G_{R\rightarrow S}(X_R)$.

%The model uses as two generators, responsible of mapping an image from the source domain to the target domain and back. The input image $X_S$, is inputed in the generator $G_{S\rightarrow T}$ which translates it into $\hat{X}_T$. The second generator $G_{T\rightarrow S}$ takes an image $X_T$ and translates it into $\hat{X}_S$. Two discriminators are responsible for distinguishing the real image from the fake one created by the generator in each domain. Discriminator $D_T$, aims at distinguishing $X_T$ from $G_{S\rightarrow R}(X_S)$ while the discriminator $D_S$ aims at distinguishing between $X_S$ and $G_{T\rightarrow S}(X_T)$.

%\subsection{Networks}

% \begin{figure}[b]
% \centerline{\includegraphics[scale=0.4]{Figures/enocderDecoder.png}}
% \caption{The encoder downsamples a tactile image by using a stride of two, whereas the encoder upsamples an image by utilising a transposed convolution of a stride of two.}
% \label{figEncoderDecoder}
% \end{figure}

% \begin{figure}[htbp]
% \centerline{\includegraphics[scale=0.53]{Figures/Generator.png}}
% \caption{The Generator of the model encodes the image into a representative feature space which is then upsampled to form the mapped image. The dotted lines represent the skip connections.}
% \label{figGenerator}
% \end{figure}

\begin{figure*}[tt]
\centerline{\includegraphics[scale=0.92]{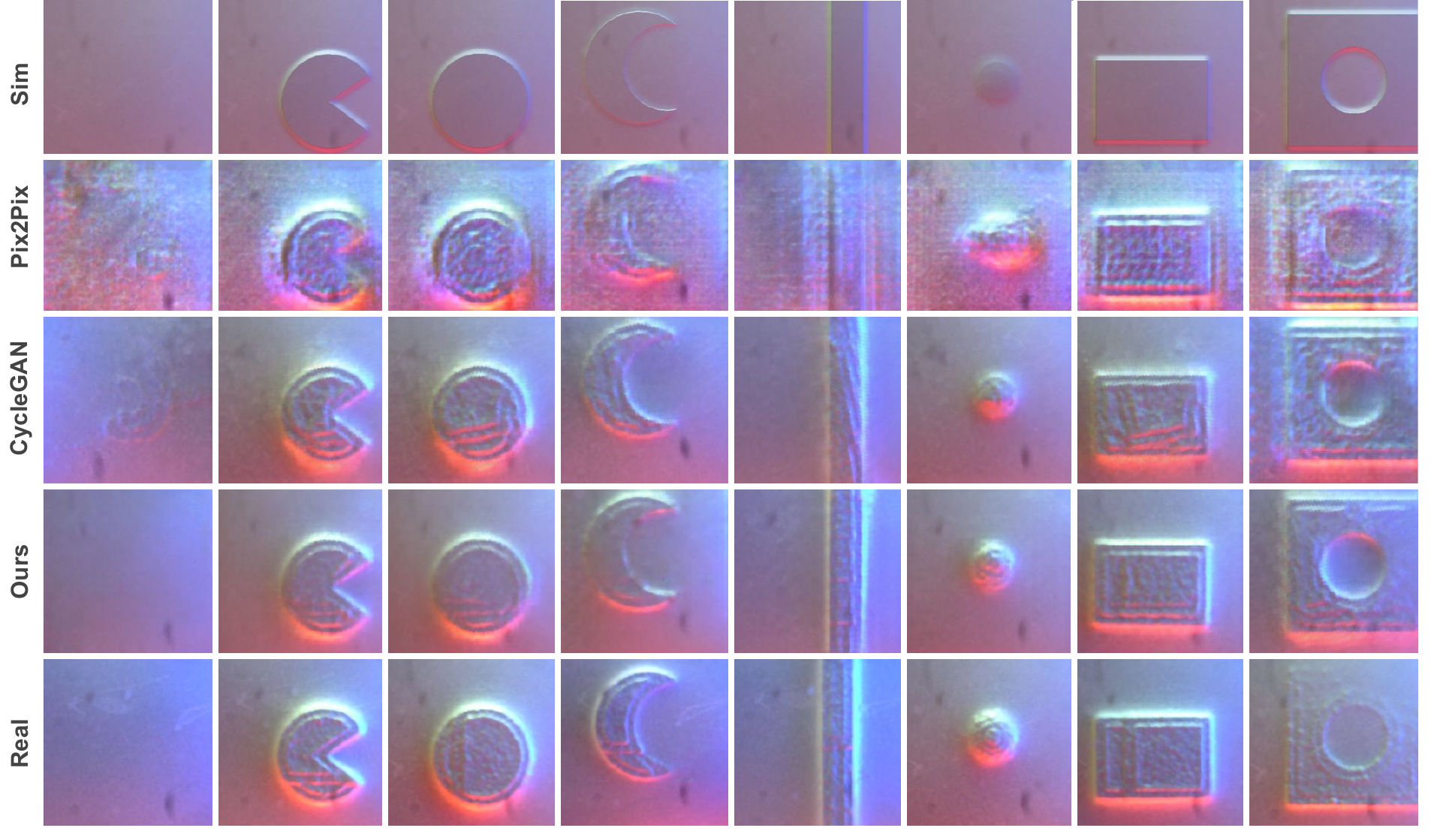}}
\caption{Top row: Simulated samples collected using the GelSight simulation approach \cite{gomes2021generation}; Bottom row: The corresponding real samples captured using a real GelSight sensor \cite{dong2017improved}. In between, second to fourth rows: The two baselines that we experiment with and our final proposed network. As seen in the listed images, the original simulation tactile images lack the textures produced by the 3D printed process that can be observed in the real samples. On the opposite extreme, Pix2Pix~\cite{pix2pix} renders over textured tactile images, including outside the in-contact areas. CycleGAN~\cite{cycleGAN} produces much cleaner textures, when compared to Pix2Pix, however, some texture leaking can still be observed, e.g., in the first and last samples (columns). Finally, our proposed network generates the best results, with the textures generated only within the in-contact areas.}
\label{figGenerativeModels}
\end{figure*}

% Generator
The generator follows a U-Net architecture~\cite{unetBasic} which consists an encoder (downsample) part and a decoder (upsample) part. Each layer in the encoder, consists of blocks of a convolutional layer, followed by an instance normalisation layer and a Leaky ReLU activation. Each convolution has a stride of two and with each layer, the number of filters is doubled until the image is reduced to height and width of one and $512$ filters. The decoder is constructed of layers that consist of a transposed convolution of stride two, followed by an instance normalisation, a dropout layer, and a ReLU activation. The dropout layer is applied to the upsample section of the generator and functions as a network regulator. The network is compelled to learn meaningful representations from the latent space as a result of this. In addition, skip connections are tied between the mirrored layers in the encoder and decoder part of the model, which allows the model to propagate context information to higher resolution layers~\cite{unetBasic}. This is done by concatenating the mirrored layer with the output of the downsample layer. As a result, each layer in the decoder has the amount of filters doubled than the corresponding mirrored layer in the decoder network. The first downsample block does not use the normalisation, while in the decoder part only the first three blocks use a dropout layer. 

% Discriminator
The Discriminator follows a Patch-GAN architecture~\cite{pix2pix}, where each layer consists of a convolution of a stride of two followed by an instance normalisation layer and a Leaky ReLU activation. The discriminator, rather than giving an absolute value, it outputs a patch of $N{\times}N$ dimensions, in our case, $N=33$. This allows the computation of $L1$ loss between the patches output by the discriminator. 

% Classification Network

\subsection{Loss Functions}

\textbf{Adversarial Loss}.  The generator $G_{S \rightarrow R}$ represents a mapping function that takes an element from the distribution $X_S$ and maps it to the distribution $X_R$ while $D_R(x_s)$ outputs the probability that an instance comes from $X_R$ rather than from $X_S$. The discriminator tries to maximise the probability of correctly assigning a label to the $X_R$ and $G_{S \rightarrow R}(X_S)$ while the generator aims to minimise $log(1-D(G_{S \rightarrow R}(X_S))$. In other words, the loss can be described as a minimax game where the generator $G_{S \rightarrow R}$ aims at translating a tactile image from the synthetic domain to the reality domain whereas the discriminator $D_R$ aims at distinguishing between a generated tactile image and a real one. This corresponds to:

\begin{equation}
\begin{aligned}
    \mathcal{L}_{GAN}(G_{S \rightarrow R},D_R,X_R,X_S) = & E_{x_r\sim X_R}[logD_R(x_r)] + & \\
    & E_{x_s\sim X_S}[log(1- & \\ 
    & D_R(G_{S\rightarrow R}(x_s))]\label{adversarialLoss1}
\end{aligned}
\end{equation}

\noindent In addition, the generator $G_{R \rightarrow S}$ learns to map the images from $X_R$ to $X_S$ while the discriminator $D_S$ distinguishes between them. This results into:

\begin{equation}
\begin{aligned}
    \mathcal{L}_{GAN}(G_{R \rightarrow S},D_S,X_S,X_R) = & E_{x_s\sim X_S}[logD_S(x_s)] + & \\
    & E_{x_r\sim X_R}[log(1-& \\
    & D_S(G_{R\rightarrow S}(x_r))]\label{adversarialLoss2}
\end{aligned}
\end{equation}

Together, \eqref{adversarialLoss1} and \eqref{adversarialLoss2} give the total adversarial loss of:

\begin{equation}
\begin{aligned}
    \mathcal{L}_{GAN} = & \mathcal{L}_{GAN}(G_{S \rightarrow R},D_R,X_R,X_S) + \\
    & \mathcal{L}_{GAN}(G_{R \rightarrow S},D_S,X_S,X_R)\label{adversarialLossCombined}
\end{aligned}
\end{equation}

\textbf{The Cycle Consistency loss}. While the tactile image generated by $G_{S \rightarrow R}(X_S)$ may learn to produce convincing results that seem like they are sampled from the real distribution, it may not preserve the information in $X_S$, such as the class and the location of the object in the image. In order to enforce the stability and consistency of the model, the cycle consistency loss ${L}_{cycle}$~\cite{cycleGAN} has been implemented. ${L}_{cycle}$ calculates the difference between the input simulation tactile image $x_{s}$ and the image translated to a real image through the generator $G_{S \rightarrow R}$ then back to the synthetic domain through the generator $G_{R \rightarrow S}$. This allows the model to learn the mappings between the domains without the need of paired data such as in CycleGAN~\cite{cycleGAN}, DualGAN~\cite{dualGAN}, and DiscoGAN~\cite{discoGAN}. Mathematically, given an image $x_s$, and the cycled image $G_{R \rightarrow S}(G_{S \rightarrow R}(x_s))$, we want $G_{R \rightarrow S}(G_{S \rightarrow R}(x_s))\approx x_s$. Similarly, for an image $x_r$, we want $G_{S \rightarrow R}(G_{R \rightarrow S}(x_r))\approx x_r$. Both of the losses give the total cycle consistency loss:

\begin{equation}
\begin{aligned}
    \mathcal{L}_{cycle}&(X_S,X_R,G_{S \rightarrow R},G_{R \rightarrow S})=\\ & \mathbb{E}_{x_s \sim X_S}[||x_s - G_{R \rightarrow S}(G_{S \rightarrow R}(x_s))||_1] +\\ 
    &\mathbb{E}_{x_r \sim X_R}[||x_r - G_{S \rightarrow R}(G_{R \rightarrow S}(x_r))||_1] \label{cycleLoss}
\end{aligned}
\end{equation}

\textbf{Identity Loss.} In order to preserve the colours when the tactile image gets translated from one domain to the other, an identity loss is introduced such that, when a simulation tactile image from $X_S$ is translated through the generator $G_{R \rightarrow S}$, the output $x_s$ should have similar colour settings from the light configurations in simulation. This results in:

\begin{equation}
    \begin{aligned}
        \mathcal{L}_{identity}= & \mathbb{E}_{x_s \sim X_S}[||x_s - G_{R \rightarrow S}(x_s)||_1] + \\
        & \mathbb{E}_{x_r \sim X_R}[||x_r - G_{S \rightarrow R}(x_r)||_1]\label{idLoss}
    \end{aligned}
\end{equation}

\textbf{Mask Loss.}  As shown in Fig.~\ref{framework}, using the depth maps in the simulation, we can distinguish between the foreground and the background by setting any region that is less than the height of the elastomer to one and the rest to zero and thus we created the binary mask $m_s$ of the object $x_s$. In order to make the areas that are not in contact unaffected by the textures, we constrain the image background on both the simulated and the real image background thus not only giving the model stability to the outside the contact regions, but also accounting for class shift that the model is prone to~\cite{cyCADA}. Furthermore, we propose to use a hyperparameter $\alpha$ to balance the background target (simulated and real) that enable us to control the generated background to copy more features from the real or simulated backgrounds. This results in the formulation of our mask loss:

\begin{equation}
    \begin{aligned}
        \mathcal{L}_{mask} & (M_S,X_S,X_R,G_{S \rightarrow R})= \\ 
        & E_{x_s\sim X_S} [\alpha ||(G_{S \rightarrow R}(x_s)-x_s)(1-m_s)||_1+\\
        &(1-\alpha)||(G_{S \rightarrow R}(x_s)-x_r)(1-m_s)||_1]
        \label{maskLoss}
    \end{aligned}
\end{equation}
where a higher $\alpha$ would mean that the tactile image is more constrained on the simulated dataset whereas a lower $\alpha$ would imply that the image is more constrained on the real dataset.

\section{The Dataset and Experiment setup}
 To carry out experiments and evaluation we make use of the dataset captured in~\cite{gomes2021generation}. This dataset consists of paired sets of simulated tactile images $X_S$, real tactile images $X_R$ and raw close-up depth maps that are collected by tapping a GelSight sensor~\cite{dong2017improved} against 21 reference objects of different shapes. These objects were modelled in CAD and printed using a Formlabs~Form~2~3D~printer. To ensure a controlled position of the sensor relative to the object, a Fused Deposition Modeling (FDM) 3D printer \textit{A30} from Geeetech was used as a Cartesian actuator, to move the sensor and tap the reference objects in $3 \times 3$ grid and 11 depths. This results in each set containing 2,079 ($ 21 \times 99 $) tactile samples. Identical setups were created both in the real world and simulation (in Gazebo), and the Robot Operating System (ROS) was used to orchestrate the different software components and the overall data collection. While in the real setup the tactile images $X_R$ were directly captured, in the simulated counterpart the close-up depth maps were firstly captured online, and then the tactile images $X_S$ were generated using the simulation method~\cite{gomes2021generation} offline. For more details of the dataset, we refer the reader to~\cite{gomes2021generation} and the project website\footnote{https://danfergo.github.io/gelsight-simulation/}.

 Despite the high resolution of the 3D printer, textures were introduced during the printing process that significantly affect the \textit{Sim2Real} transfer. For instance, in Fig.~\ref{figGenerativeModels} it can be observed that the real samples present different textures compared to the ones in the simulated counterparts. Furthermore, it can be seen that the difference between the real and simulated samples are in the high frequency texture, while the overall shapes of the model are the same. Even though this texture could be further smoothed using a variety of methods, we keep them and consider them as unexpected artefacts that could result from natural and unpredictable wear of the object that are commonly seen in the real life. 

%1. Preprocessing the data \\
In order to conduct our experiments we first preprocessed the data. For the training dataset, we first normalised the tactile images at the pixel level into the $[-1;1]$ interval. We then employed a data augmentation method, in which we increased the resolution of the tactile images and applied a random crop over the tactile images, followed by a slight rotation and a horizontal flip applied randomly.
%2. Implementation details \\
We implemented all of the models using the Keras API available through Tensorflow.

 \begin{table}[t]
\caption{Classification Task Summary}
\def\arraystretch{1.2}
\begin{center}
\begin{tabular}{c|c|c}
\hline

 \textbf{\textit{Model}}& \textbf{\textit{Sim}}& \textbf{\textit{Real}}\\
\hline
Direct & \textbf{91.90}\% $\pm 1.80$ & $53.47\%\pm 6.64$ \\
Pix2Pix & $91.07\%\pm 0.95$ & $60.53\%\pm 2.81$\\
CycleGAN & $90.07\%\pm 1.04$ & $85.57\%\pm 3.36$ \\
CycleGAN w Mask Sim & $91.07\%\pm 1.15$ & \textbf{90.26} \%$\pm 2.70$\\
CycleGAN w Mask Real & $90.41\%\pm 0.89$ & $86.25\%\pm 5.15$ \\
CycleGAN w Mask Combined & $90.55\%\pm 0.84$ & $89.17\%\pm 2.45$ \\
\hline
%\multicolumn{4}{l}{$^{\mathrm{a}}$Sample of a Table footnote.}
\end{tabular}
\label{tab2}
\end{center}
\end{table}

\section{Experiments and discussion}

% Overview
%We consider the task of domain adaptation where the labels of the test set are unknown. The aim is to train a classifier on the source dataset $X_S$ so that it can generalise on the target dataset $X_R$ without a significant drop in performance. We evaluate our methods both quantitatively and qualitatively. For the qualitative aspect, the results can be observed in Fig.~\ref{figGenerativeModels}.

%First, we follow the naive approach of training a model on the source domain dataset $X_S$ and testing its performance on the target domain $X_R$. For this purpose, we use the ResNet50 architecture \cite{resNet} with the weights pretrained on the "ImageNet" dataset.  On top of the base model, two blocks composed of a dense layer, batch normalisation and an ELU activation were added. Additionally, we add an output layer composed of 21 neurons and a softmax activation.

%Next we considered the dataset as a paired one and we trained the generator in an adversarial fashion where we trained the discriminator $D_R$, to distinguish the generated image $G_{S \rightarrow R}(x_s)$ from it's pair from $X_R$. We add the generator as the input layer to the classification model thus training the classification model on the mapped images $G_{S \rightarrow R}$. We next adopt an unsupervised approach and train the model using \eqref{cycleLoss} and \eqref{idLoss} followed by the implementation of \eqref{maskLoss}. 

We  evaluate  the  proposed texture generation network with  three  sets  of experiments. Firstly, we compare the generated tactile images against corresponding adapted samples, with both quantitative and qualitative analyses; and then, we demonstrate the advantages of considering the adapted, instead of the original simulated images, for Sim2Real transfer learning in a classification task. As shown in Fig.~\ref{figGenerativeModels}, the generated tactile images using our proposed network appear substantially more similar to the real images than the simulated counterparts, and from Table~\ref{tab2} it can be seen that the initial drop in performance  caused by the Sim2Real gap of $38.43\%$ is reduced to $0.81\%$ when considering the images adapted by our network.

\begin{table}[b]
\caption{Real and Adapted comparison}
\def\arraystretch{1.2}
\begin{center}
\begin{tabular}{c|c|c}
\hline
 \textbf{\textit{Model}}& \textbf{\textit{SSIM $\uparrow$}}& \textbf{\textit{MAE $\downarrow$}}\\
\hline
Pix2Pix & $0.332$ & $30.80\%$ \\
CycleGAN & $0.631$ & $23.26\%$ \\
CycleGAN w Mask Sim & $0.734$ & $10.70\%$ \\
CycleGAN w Mask Real & \textbf{0.751} & $10.80\%$ \\
CycleGAN w Mask Combined& $0.719$ & \textbf{10.50\%} \\
\hline
%\multicolumn{4}{l}{$^{\mathrm{a}}$Sample of a Table footnote.}
\end{tabular}
\label{tab1}
\end{center}
\end{table}

\subsection{Constraining the augmented texture areas}

%1. Overview \\
During our early experimentation phase, when analysing tactile images generated CycleGAN models~\cite{cycleGAN} we observe that they produce realistic results with only slight discrepancies from the real tactile images. 
%2. Problem \\
However, one tendency of the CycleGAN is to mirror the background and light of the tactile images. Furthermore, the model is not constrained to maintain the object structure while being mapped~\cite{cyCADA}. While this result is not entirely detrimental for cases such as classification, where one can associate such behaviour with a domain randomisation technique, it highlights the instability of the model. Such instability can be observed in the column one, of the CycleGAN row, of Figure~\ref{figGenerativeModels}, where the background is flipped and anomalies are injected into the picture thus creating deformations. 
%3. Solution \\
To mitigate the issue, we constrain the model on the background of both the simulated and the real pair of image by using our proposed mask loss in Eq.~\eqref{maskLoss}. We further weight the terms differently in relation to the background of provenience such as, a weight of $0.4$ implies that the latter would be multiplied by a weight of $0.6$ and giving a total error based on both backgrounds. We test both of the extremes, constraining the model on only simulation background, only real background as well as the mixture of two. With the mask loss implemented, we observed a greater stability of the background, where the flip of colours along with the background does not occur. Furthermore, the model applies different textures on the contact zones, adding scratches at different angles and on different figures or not adopting a particular scratch. This has the potential to minimise the situation where the model runs into an unexpected type of scratch.

\begin{figure}
\centerline{\includegraphics[scale=0.15]{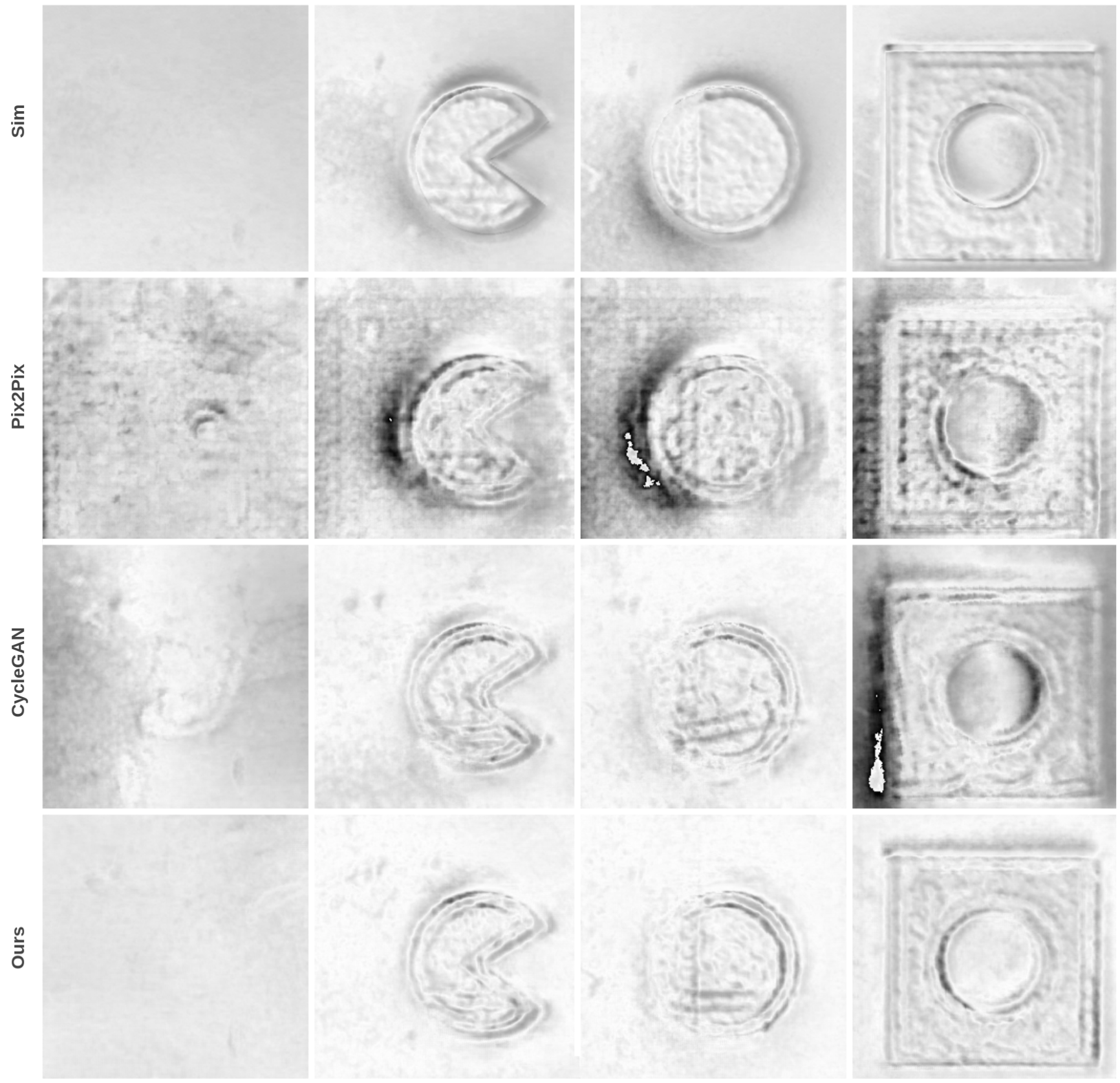}}
\caption{Difference maps of the generated adapted tactile images, using the different studied methods, against the real reference, with white pixels representing zero difference. As seen in the figure, the textures of the real images are directly visible in \textbf{Sim}, demonstrating the smoothness of the original simulated images. \textbf{Pix2Pix}~\cite{pix2pix} produces randomised textures throughout the entire image, resulting in significant differences even in areas of no contact. \textbf{CycleGAN}~\cite{cycleGAN} produces better results than~\textbf{Pix2Pix}, however, some artefacts can be seen in areas of non contact, e.g., first column, and a distortion on the pose of the object is visible in the last column. Finally, \textbf{Ours} method produces the overall smaller differences. }
\label{diffs}
\end{figure}

\subsection{Comparison of different domain adaptation methods}
% %1. SSIM overview \\
% Firstly, we distinguish the mapping methods by using the Structure Similarity Index Metric (SSIM).  The SSIM evaluates an image by using a combination of metrics that aim at mimicking the visual perception of humans. As a result, the metric compares the luminance, contrast and the structure between two pictures \cite{SSIM}. 
%2. SSIM implementation \\
In order to compare different domain adaptation methods quantitatively, we compute the average Structural Similarity (SSIM) and Mean Absolute Error (MAE) between the adapted images generated by the different methods and the real corresponding pairs. The obtained results are reported in Table~\ref{tab1}.
%3. MSE \\
We further compute the relative absolute differences maps, between the samples generated by each method and real counterparts, to improve the understanding of the numerical results, shown in Fig.~\ref{diffs}.
%4. Results \\
Our method of adding information from the background results in the greatest SSIM score $0.751$ when being constrained on the real background, while managing to achieve the lowest MAE ($0.105$) when using a mixed background approach. The Pix2Pix network ~\cite{pix2pix}, although it can create realistic samples, requires a greater amount of time to converge and the model is free to shift the location of the objects freely, resulting in a lower value on SSIM. Furthermore, the random light flipping that we observe will affect the value negatively.

\subsection{Sim2Real transfer for object classification}

%1. Overview
To evaluate the advantages of considering the adapted tactile images \textit{versus} the original simulations for Sim2Real learning, we consider a simple task of object classification using tactile images. To this end, we start by mapping all the simulated images to the target domain, using the pre-trained CycleGAN on which we further add our structural constraint, and proceed by training a classification model on the mapped images. For this purpose, we use the ResNet50 architecture~\cite{resNet} with the weights pretrained on the ImageNet dataset. On top of the base model, two blocks composed of a dense layer, batch normalisation and an ELU activation were added. For each of the added layers, we use the He initialisation \cite{initialization:he} to avoid the problem of vanishing and exploding gradients present in deep architectures. In addition, we add an output layer composed of 21 neurons and a softmax activation. We repeat the procedure of training the classifier and testing the results for 10 times. Each time we train the classifier for 30 epochs. We then test the models on the target domain by computing the accuracy of the model. The results are presented in table Table~\ref{tab2}.
%2. Short analysis of the results \\
The direct transfer between the two domains shows the greatest amount of gap, with a drop of $38.43\%$ whereas our method has the least amount of drop ($0.81\%$) when the mask loss relies mostly on the simulated background as well as the greatest accuracy on the testing dataset ($90.26\%$). 
%3. Conclusion of the result \\
The results show that by providing the model with background information, the model is more stable and does not shift the classes, which has been encountered in previous works~\cite{yang2018unpaired,cycleGAN}.

\section{Conclusion}

In this paper, we proposed a novel texture generation network that is capable of bridging the gap between simulation and reality in the context of tactile images generated with a GelSight sensor. This allows the convenient training of other models in a simulated environment thus reducing the cost and the damage that can occur if the model is transferred to the domain of reality directly. Besides the ability to bridge the gap, the model is capable of generating new textures on the same object thus acting as a domain randomised and increasing the robustness of a model that is trained in simulation. We discovered that anomalies are created in the differences between the contact areas and the background and we stabilised the model using a proposed mask loss. In the future work, we would like to implement our proposed method on more complex Sim2Real tasks, for example, robot grasping and manipulation with tactile sensing.

\bibliography{Bibliography.bib}{}
\bibliographystyle{ieeetr}
\vspace{12pt}

\newpage

\onecolumn

\end{document}